\documentclass[runningheads]{llncs}

\usepackage[T1]{fontenc}
\usepackage{graphicx}
\usepackage{amssymb}
\usepackage{amsmath}
\usepackage{enumitem}
\usepackage{subcaption}
\usepackage{multirow}
\usepackage{color}

\begin{document}
\title{Comparison of Clustering Algorithms for Statistical Features of Vibration Data Sets}
\titlerunning{Comp. of Clustering Alg. for Stat. Features of Vibration Data Sets}
\author{Philipp Sepin\inst{1,2} \and
Jana Kemnitz\inst{1} \and
Safoura Rezapour Lakani\inst{1} \and
Daniel Schall\inst{1}}
\authorrunning{P. Sepin et al.}
\institute{Siemens Technology \and
Vienna University of Technology}
\maketitle
\begin{abstract}
Vibration-based condition monitoring systems are receiving increasing attention due to their ability to accurately identify different conditions by capturing dynamic features over a broad frequency range. However, there is little research on clustering approaches in vibration data and the resulting solutions are often optimized for a single data set. 
In this work, we present an extensive comparison of the clustering algorithms K-means clustering, OPTICS, and Gaussian mixture model clustering (GMM) applied to statistical features extracted from the time and frequency domains of vibration data sets. Furthermore, we investigate the influence of feature combinations, feature selection using principal component analysis (PCA), and the specified number of clusters on the performance of the clustering algorithms. We conducted this comparison in terms of a grid search using three different benchmark data sets.
Our work showed that averaging (Mean, Median) and variance-based features (Standard Deviation, Interquartile Range) performed significantly better than shape-based features (Skewness, Kurtosis). In addition, K-means outperformed GMM slightly for these data sets, whereas OPTICS performed significantly worse.
We were also able to show that feature combinations as well as PCA feature selection did not result in any significant performance improvements. With an increase in the specified number of clusters, clustering algorithms performed better, although there were some specific algorithmic restrictions.

\end{abstract}
\section{Introduction}
The constant and accurate monitoring of machinery is a vital aspect of its operation. Vibration-based condition monitoring systems are receiving increasing attention due to their ability to accurately identify different conditions by capturing dynamic features over a broad frequency range \cite{Ruiz2014,Panda2018,Elangovan2011,Romero2018,Ribero2020,Kolar2020,Jafarian2018,Venkata2019,Vos2022}. Further, low-cost sensors enable large scale operation through various equipment types \cite{Heistracher2022}. In this context, unsupervised learning methods can prove instrumental as a preprocessing step for supervised learning methods or as a stand-alone method when dealing with missing labels. Research in the field mainly focused on classification \cite{Holly2022,Kemnitz2022,Heistracher2021,Ruiz2014,Panda2018,Elangovan2011,Altobi2019,Romero2018,Ribero2020,Kolar2020,Jafarian2018,Zhao2021,Zabihi-Hesari2019,Venkata2019} or anomaly detection \cite{Heistracher2022,Vos2022,Baraldi2015}.

What might seem trivial as a supervised classification task bears significant difficulties when done in an unsupervised way. K-means clustering and DBSCAN (density-based spatial clustering of applications with noise) \cite{Ester1996} have been explored for condition classification of bearings using vibrational data \cite{Kerroumi2013}. While DBSCAN is very sensitive to clustering parameters and has difficulty detecting clusters of different densities, the extension OPTICS (ordering points to identify the clustering structure) \cite{Ankerst1999} was shown to solve these issues when used for condition classification of bearings \cite{Hotait2021}. Fuzzy C-means clustering (FCM) has been utilized for detecting anomalous conditions of nuclear turbines \cite{Baraldi2015}.

Besides the selection of the right clustering algorithms, feature extraction and selection itself is a challenging task in vibrational data since the performance of the clustering algorithm heavily depends on the features. Previous work explored statistical time domain features \cite{Ruiz2014,Obuchowski2016,Dhamande2018} and frequency domain features extracted by means of the fast Fourier transform (FFT) \cite{Zabihi-Hesari2019,Jafarian2018}, discrete wavelet transform (DWT) \cite{Jafarian2018,Zabihi-Hesari2019,Heistracher2022,Dhamande2018}, and continuous wavelet transform (CWT) \cite{Altobi2019,Dhamande2018}. The difficulty here lies in the optimal modeling of the feature space to allow for the unsupervised separation of different conditions, which is particularly difficult in the case of industrial data and gets even more difficult with an increasing number of conditions.

Summarizing, there is little research on clustering approaches in vibration data and the resulting solutions are often optimized for a single data set. A fundamental analysis of feature extraction and selection methods, and clustering algorithms validated over several data sets is required. Therefore, we aim to answer the following questions.

\begin{enumerate}[label=\textbf{Q\arabic{*}.}, topsep=1pt]
    \small
    \item Which combinations of statistical features and clustering algorithms perform best for multiple data sets?
    \item Does the performance of statistical feature and clustering algorithm combinations generalize for arbitrary data sets?
    \item Can the combination of several different features improve the performance of the clustering algorithms?
    \item Can principal component analysis (PCA) improve the performance of the clustering algorithms by selecting the most representative features?
    \item How does the specified number of clusters affect the performance of the clustering algorithms?
\end{enumerate}

\section{Theoretical Foundations}
\subsection{Clustering}
The K-means algorithm is one of the most popular iterative clustering methods. One chooses the desired number of cluster centers and the K-means algorithm iteratively moves the centers to minimize the total within cluster variance \cite{elementsofstatlearn}.

Gaussian mixture model clustering (GMM) can be thought of as a method similar in spirit to K-means. Each cluster is described in terms of a normal distribution, which has a centroid as in K-means \cite{elementsofstatlearn}.

DBSCAN is a density-based clustering algorithm that works by differentiating between low-point-density regions and high-point-density regions. The points are assigned to one of three categories using two density parameters. The different clusters are formed by core and border points \cite{Ester1996}. Due to DBSCAN using a global density parameter, it is not possible to reliably detect clusters with significantly different densities. To solve this, several different density parameters would be needed. This is done by the clustering algorithm OPTICS, which works in principle like an extended DBSCAN algorithm for an infinite number of distance parameters, which are smaller than a global distance parameter, which may even be set to infinity \cite{Ankerst1999}.

\subsection{Statistical Features and Principal Component Analysis}
Statistical features can be obtained from the time domain (denoted by \textit{TD}), as well as the frequency domain (denoted by \textit{FD}) by means of fast Fourier transforms (FFT) \cite{spectralanalysis,digitalsignal}. In the time domain, these measures are derived from the vibrational amplitudes, in the frequency domain, they are derived from the frequency components. This method can be enhanced by using preprocessing operations like band-pass filters for the extraction of features of specific frequency components. The following statistical features were used.
\begin{itemize}[topsep=2pt]
    \small
    \item Arithmetic mean of absolute values (\textit{Abs Mean}).
    \item Median of absolute values (\textit{Abs Median}).
    \item Standard deviation (\textit{Std}).
    \item Interquartile range (\textit{IQR}).
    \item Skewness of absolute values (\textit{Abs Skew}).
    \item Kurtosis of absolute values (\textit{Abs Kurt}).
\end{itemize}
Principal component analysis is a method for obtaining new uncorrelated variables that are linear combinations of the original variables. Due to the fact, that the principal components are sorted in order of variance in the original data, one has the option to reduce the dimensions of the input vector by only using the first few principal components, whilst still preserving most of the contained information \cite{patternrecog}. 

\section{Data Sets}
\subsection{Data Set 1}
\begin{figure}[htb]
    \centering
    \includegraphics[width=0.95\textwidth]{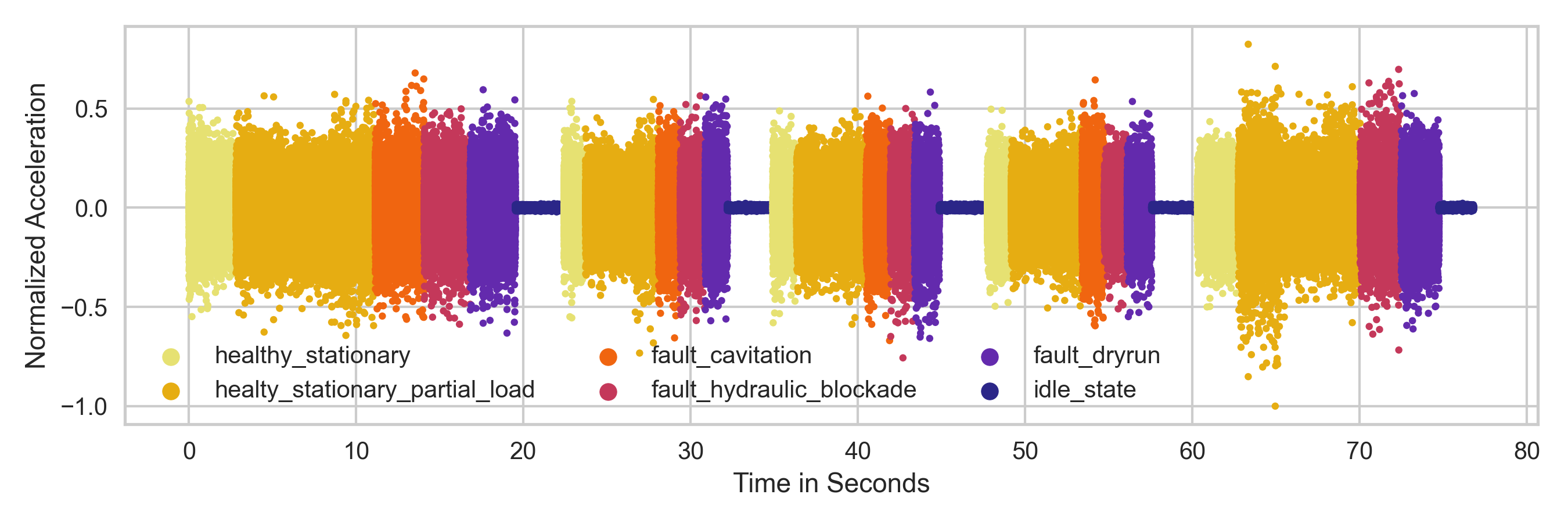}
    \caption{Time series data of data set 1 with constant components removed}
    \label{fig:Data Set 1}
\end{figure}

This data set (Fig. \ref{fig:Data Set 1}) was acquired by SIEMENS for the development of anomaly detection and classification algorithms \cite{Kemnitz2022,Holly2022}. A test bench with a centrifugal pump and a multi-sensor \cite{Bierweiler2019} was constructed for simulation of anomalous conditions. The three-axis accelerometer of the multi-sensor was used to record 512 samples at a sampling rate of 6644 Hz once every minute. This data set contains the following six conditions.
\begin{itemize}[topsep=2pt]
    \small
    \item Class 0, idle state. The system operates under normal condition.
    \item Class 1, healthy partial load. The system operates under normal condition with partial load.
    \item Class 2, healthy. The system operates under normal condition.
    \item Class 3, hydraulic blockade. The outlet valve behind the pump is closed.
    \item Class 4, dry run. The inlet valve in front of the pump is closed.
    \item Class 5, cavitation.
\end{itemize}

\subsection{Data Set 2}
\begin{figure}[htb]
    \centering
    \includegraphics[width=0.95\textwidth]{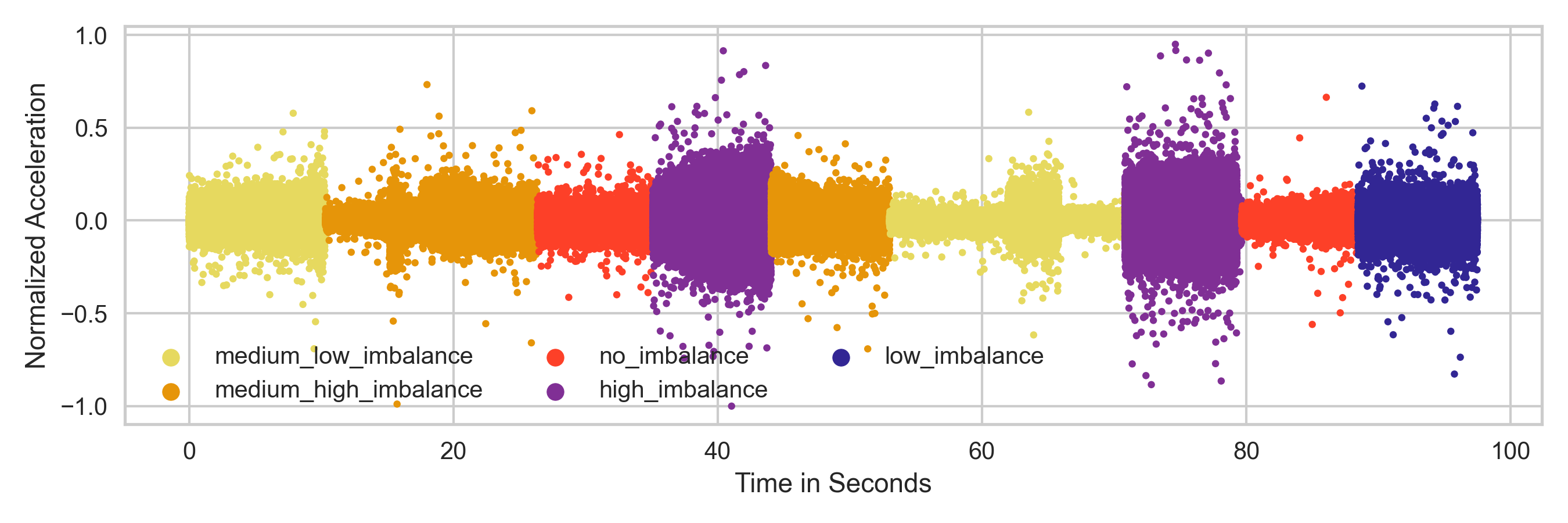}
    \caption{Time series data of data set 2 with constant components removed}
    \label{fig:Data Set 2}
\end{figure}

\noindent This open-source data set (Fig. \ref{fig:Data Set 2}) was part of a publication on the development and evaluation of algorithms for unbalance detection \cite{Mey2020}. Unbalances of various sizes were attached to a rotating DC motor shaft. Three single-axis accelerometers were used to record vibrations on the rotating shaft at a sampling rate of 4096 Hz. A statistically representative randomly shuffled subset of this data set was used. This data set contains the following five conditions.
\begin{itemize}[topsep=2pt]
    \small
    \item Class 0, no unbalance.
    \item Class 1, low unbalance.
    \item Class 2, medium low unbalance.
    \item Class 3, medium high unbalance.
    \item Class 4, high unbalance.
\end{itemize}

\subsection{Data Set 3}
\begin{figure}[htb]
    \centering
    \includegraphics[width=0.95\textwidth]{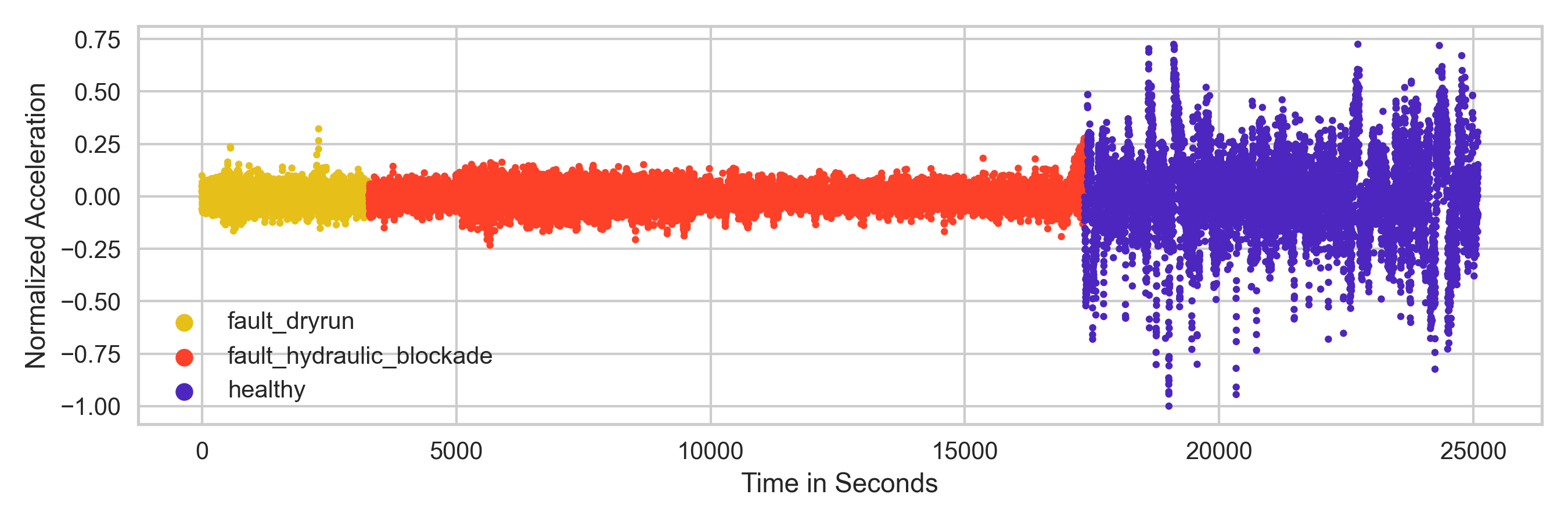}
    \caption{Time series data of data set 3 with constant components removed}
    \label{fig:Data Set 3}
\end{figure}

\noindent The Skoltech Anomaly Benchmark (SKAB) (Fig. \ref{fig:Data Set 3}) is an open-source data set designed for evaluating anomaly detection algorithms \cite{skab}. A test bench with a water circulation system was constructed for simulation of anomalous conditions. Data from two single-axis accelerometers was used. These sensors were attached to the pump and recorded vibrations at a sampling rate of 1 Hz. This data set contains the following three conditions. A fourth class that contained several different anomalous conditions was discarded in our case, since such a heavily mixed class is not suitable for unsupervised classification.
\begin{itemize}[topsep=2pt]
    \small
    \item Class 0, healthy. The system operates under normal condition.
    \item Class 1, dry run. The inlet valve in front of the pump is closed.
    \item Class 2, hydraulic blockade. The outlet valve behind the pump is closed.
\end{itemize}

\begin{figure}[htb]
    \centering
    \includegraphics[width=0.92\textwidth]{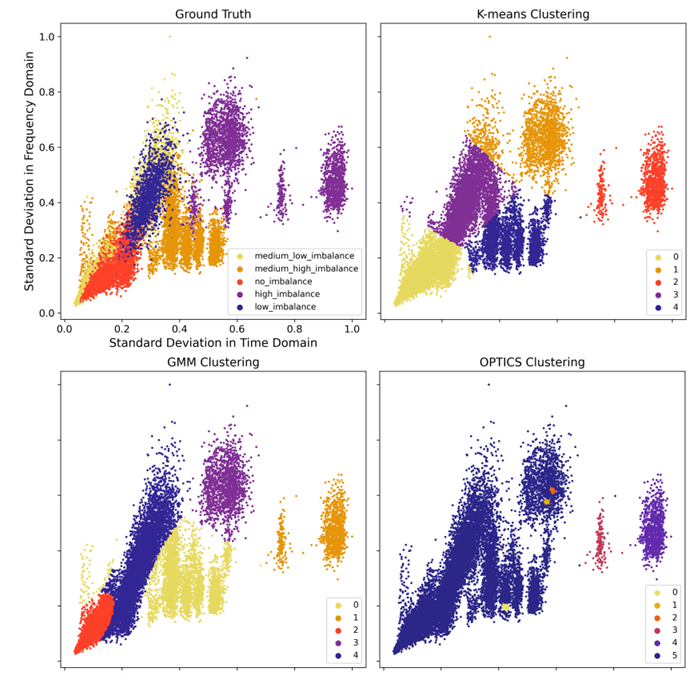}
    \caption{Feature space of data set 2 with ground truth, and clusters formed by different clustering algorithms} 
    \label{fig:FraunhoferPlotsCombined}
\end{figure}

\section{Experiments and Results}
The success of the following experiments was measured by the average purity of the resulting clusters. Purity is a measure of the degree to which clusters only contain a single class. For each cluster \(m\), the data points that belong to the class \(d\) that makes up the majority of the cluster are counted and divided by the total number \(N\) of data points. This metric does not penalize an increasing number of clusters. Therefore, it should always be seen in relation to the specified number of clusters. For each of the following experimental settings, three tests were run.
\begin{equation}
    Purity = \frac{1}{N} \sum_{m} \max_{d}|m \cap d|
\end{equation}

\subsection{Experiments}
For the following experiments \textbf{Q1} to \textbf{Q4}, the number of specified clusters was set equal to the number of conditions in the respective data set. For experiment \textbf{Q5}, the number of specified clusters was varied. Preprocessing was done by removing the constant components of the data, normalizing it, and applying a Savitzky-Golay filter \cite{Savitzky1964} with a polynomial order of \(7\) and a window size of \(9\). 

\textbf{Q1} In order to evaluate the performance of certain combinations of statistical features and clustering algorithms on the three data sets, an extensive grid search was conducted. The four variables of this grid search were the algorithm \(\in\)\{\textit{K-means, OPTICS, GMM}\}, the statistical feature \(\in\)\{\textit{Abs Mean, Abs Median, Std, IQR, Abs Skew, Abs Kurt}\}, and the domain \(\in\)\{\textit{Time Domain, Frequency Domain}\}, which resulted in a number of \(324\) trials for this evaluation.

\textbf{Q2} These results were also used to test the generalization behavior of feature algorithm combinations for the different data sets.

\textbf{Q3} For the purpose of evaluating the effect of feature combinations on the clustering performance, another grid search was conducted. For each of the algorithms, the three best performing features were chosen. For each of these sets of three features, all permutations of single, double, and triple feature combinations were tested. The three variables of this grid search were the algorithm \(\in\)\{\textit{K-means, OPTICS, GMM}\} and the feature combinations \(\in\)\{\(A\), \(B\), \(C\), \(AB\), \(BC\), \(CA\), \(ABC\)\}, which resulted in a number of \(126\) trials for this evaluation.

\textbf{Q4} In order to test the effect of feature selection using PCA on the clustering performance, another grid search was conducted. For each of the algorithms, the three best performing features were chosen and used as a feature combinations to increase feature dimensionality. The three variables of this grid search were the algorithm \(\in\)\{\textit{K-means, OPTICS, GMM}\} and the number of principal components \(\in\)\{\textit{No PCA}, \(6\), \(4\), \(2\), \(1\)\}, which resulted in a number of \(90\) trials for this evaluation. 

\textbf{Q5} A final grid search was conducted for the purpose of evaluating the effect of the specified number of clusters on the clustering performance and comparing it to the number of clusters proposed by the elbow method. For each of the algorithms, the three best performing features were chosen and used as a feature combinations. The three variables of this grid search were the algorithm \(\in\)\{\textit{K-means, OPTICS}\} and the specified number of clusters \(\in\)\{\textit{Elbow Method}, \(n\), \(1.25n\), \(1.5n\), \(1.75n\), \(2n\)\} with \(n\) as the number of conditions in the data set, which resulted in a number of \(108\) trials for this evaluation. 

\subsection{Results}
\begin{figure}[htb]
    \centering
    \includegraphics[width=0.95\textwidth]{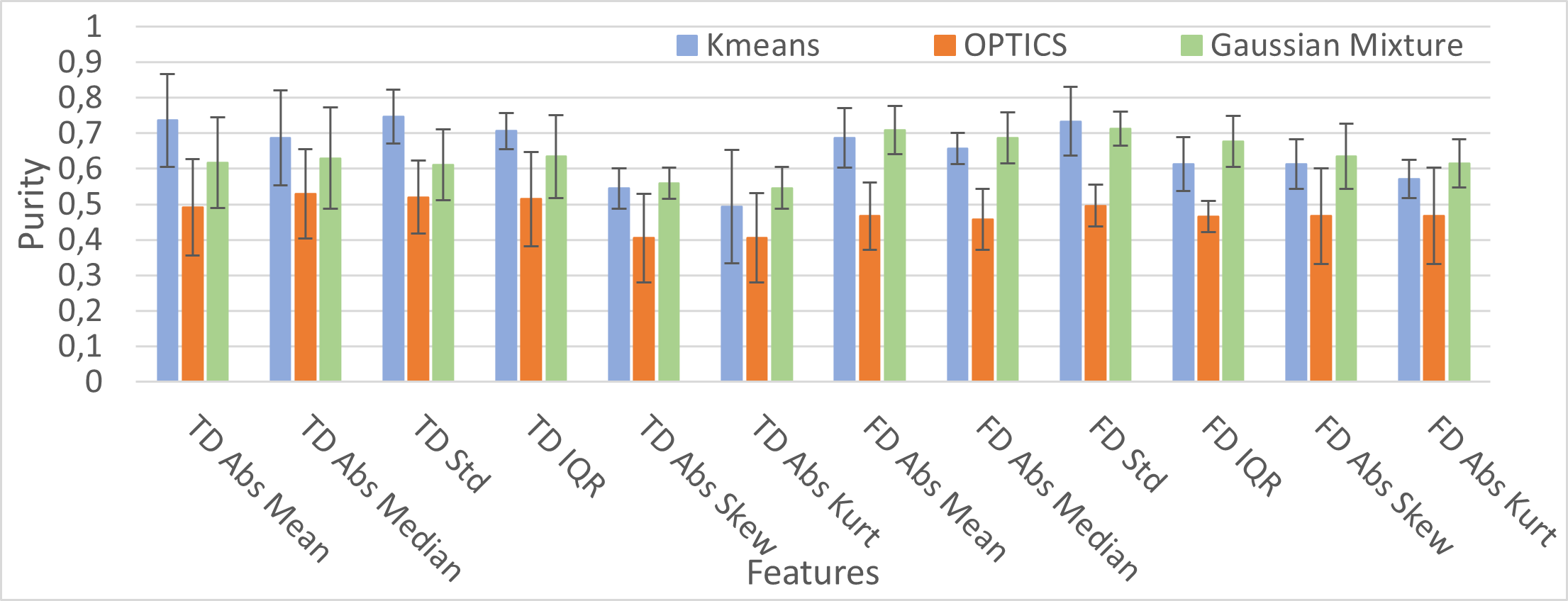}
    \caption{Average purity per feature for different clustering algorithms}
    \label{fig:GlobalPurityPerFeature}
\end{figure}

\hspace{\parindent} \textbf{Q1} Fig. \ref{fig:GlobalPurityPerFeature} shows the average purity per feature for the three different algorithms. In our case, K-means outperformed GMM slightly for these data sets, whereas OPTICS performed significantly worse than the other two algorithms.

\begin{figure}[htb]
    \centering
    \includegraphics[width=0.95\textwidth]{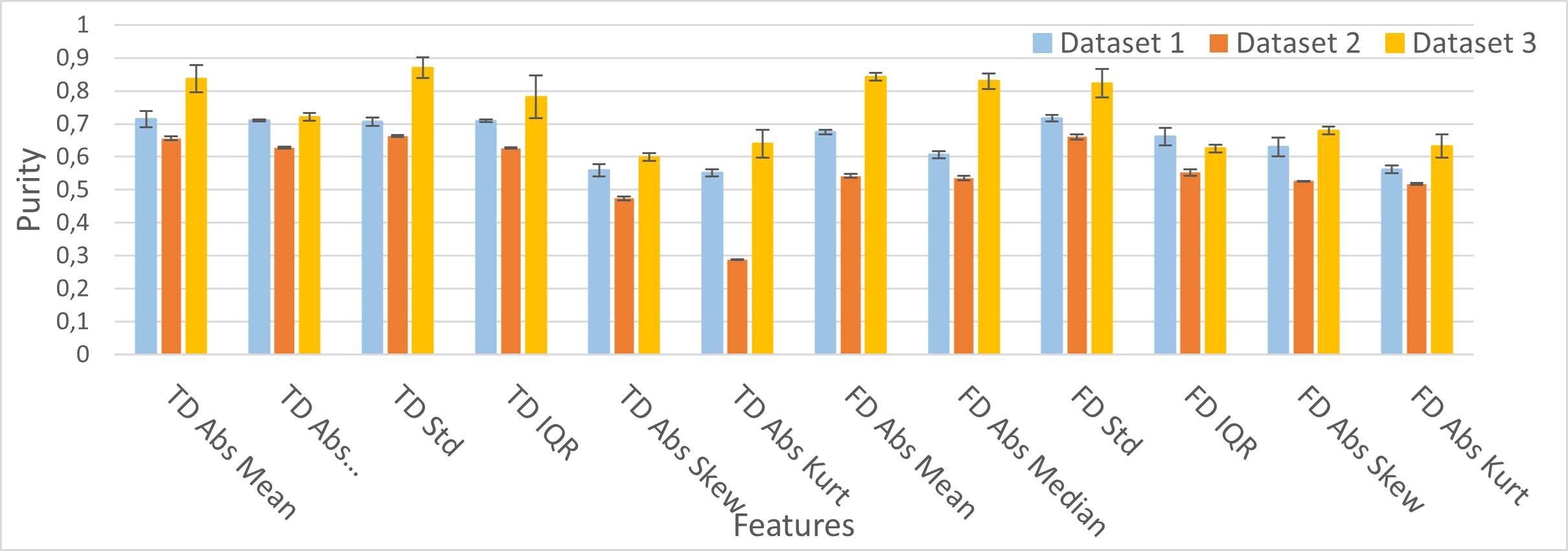}
    \caption{K-means clustering purity per feature for different data sets}
    \label{fig:KmeansPurityPerFeature}
\end{figure}

\begin{figure}[htb]
    \centering
    \includegraphics[width=0.95\textwidth]{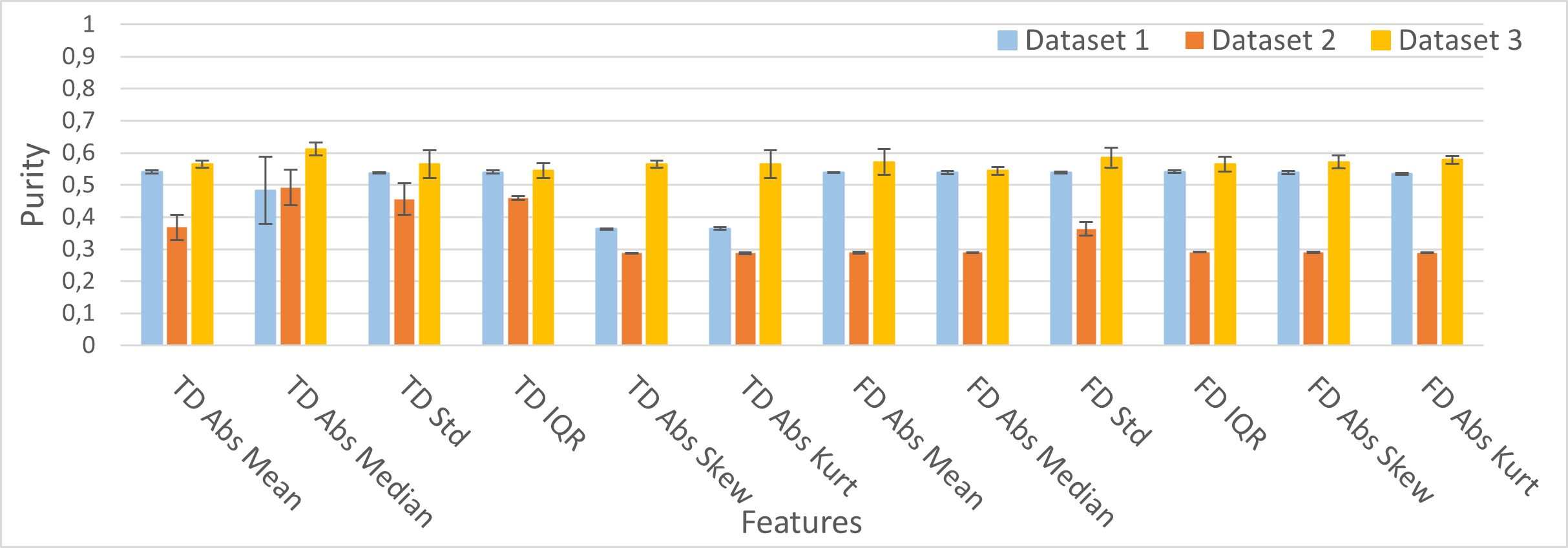}
    \caption{OPTICS purity per feature for different data sets}
    \label{fig:OPTICSPurityPerFeature}
\end{figure}

\begin{figure}[htb]
    \centering
    \includegraphics[width=0.95\textwidth]{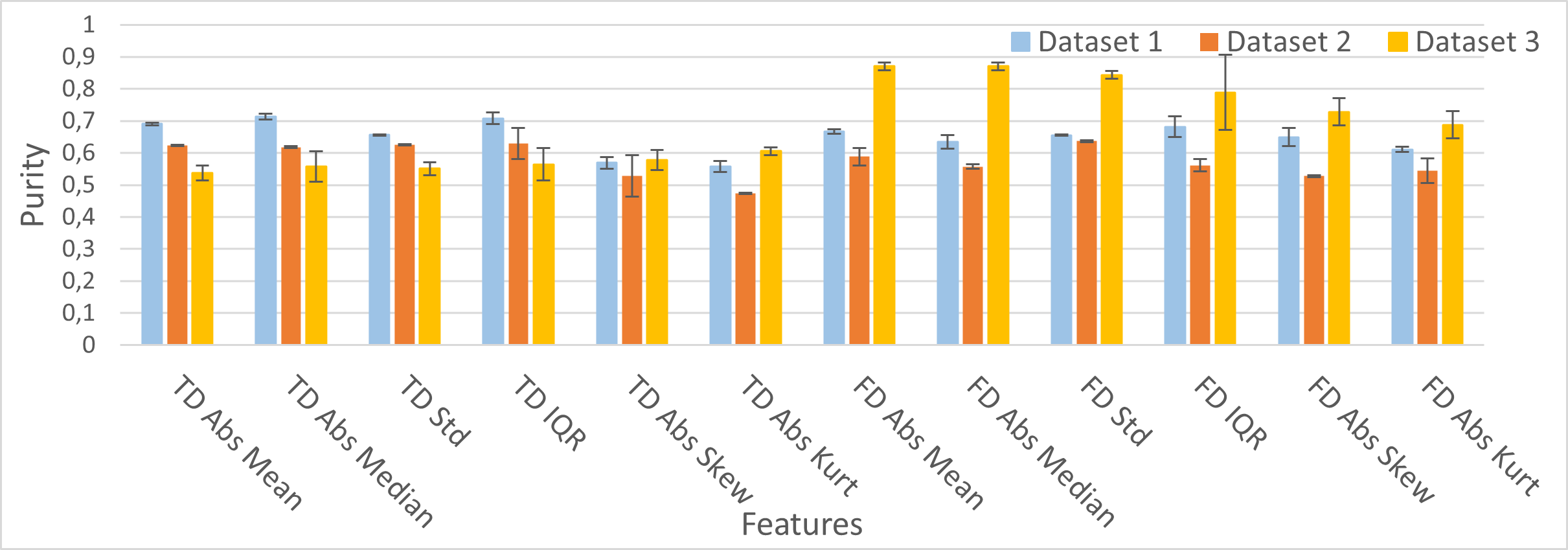}
    \caption{GMM clustering purity per feature for different data sets}
    \label{fig:GMMPurityPerFeature}
\end{figure}

\textbf{Q2} Fig. \ref{fig:KmeansPurityPerFeature} shows the purity per feature of K-means clustering for the three different data sets. Even though some features seem to be superior for all three data sets, their performance does not generalize for all these data sets. Fig. \ref{fig:OPTICSPurityPerFeature} shows the purity per feature of OPTICS for the three different data sets. As one can see, OPTICS performed significantly worse than the other two algorithms for any feature. Fig. \ref{fig:GMMPurityPerFeature} shows the purity per feature of GMM clustering for the three different data sets. For data set 3, GMM performed significantly better using features in the frequency domain than in the time domain. Nevertheless, the performance of the individual features did not generalize for all three data sets.

\begin{figure}[htb]
    \centering
    \subfloat[\centering]{{\includegraphics[width=0.498\textwidth]{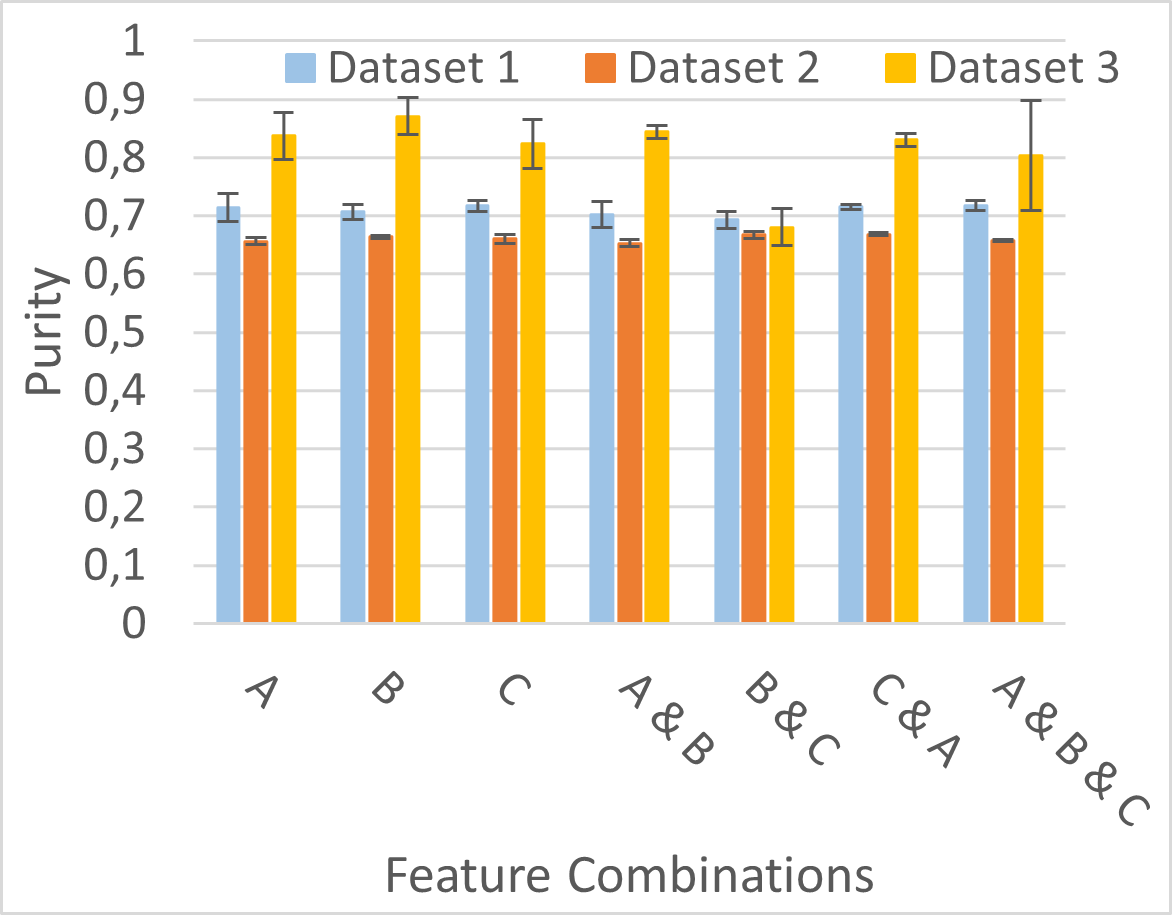}}}
    \subfloat[\centering]{{\includegraphics[width=0.502\textwidth]{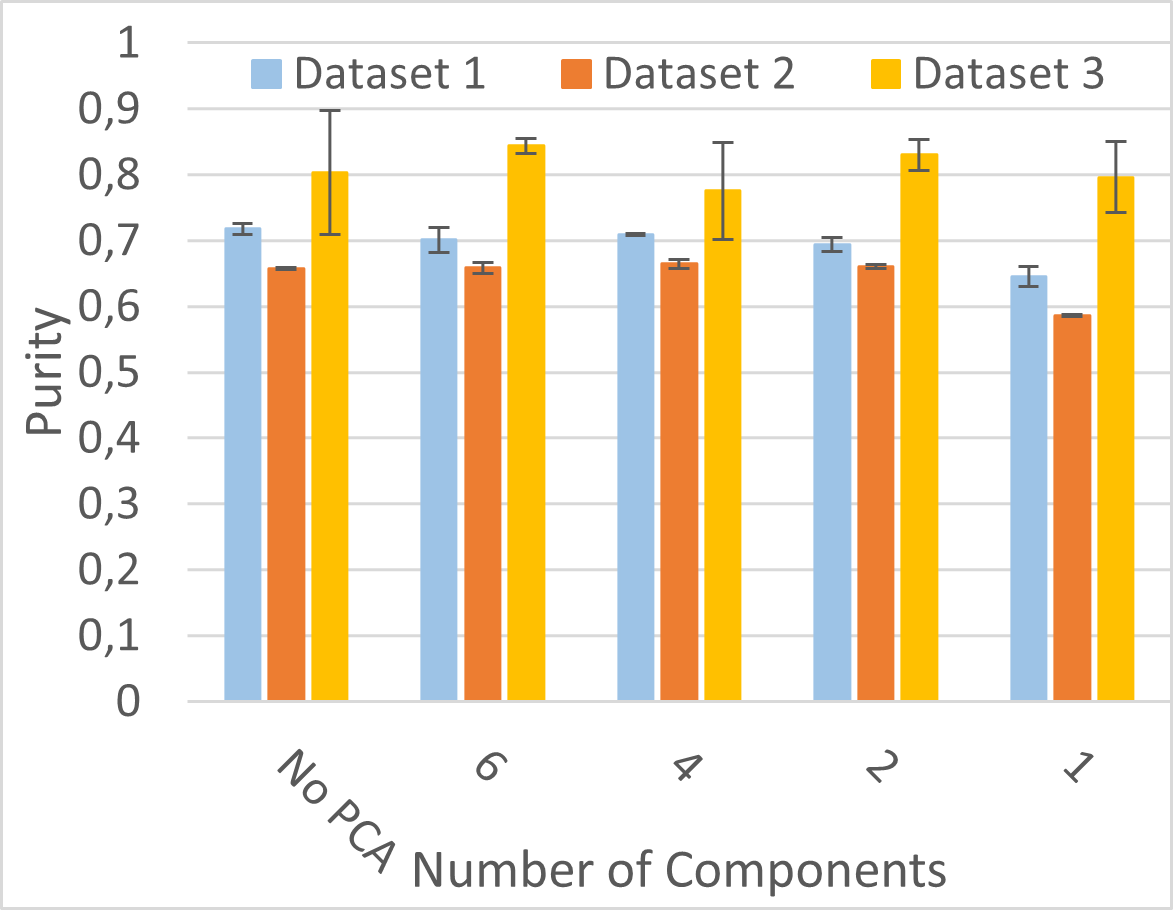}}}
    \caption{K-means clustering purity for feature combinations (a) and with different numbers of principal components (b) for different data sets}
    \label{fig:FeatureCombinationAndPCA}
\end{figure}

\textbf{Q3} Fig. \ref{fig:FeatureCombinationAndPCA} shows the purity for different feature combinations of K-means clustering for the three different data sets. Feature combinations did not significantly increase the clustering performance. The same applies to GMM clustering (data not shown).

\textbf{Q4} Fig. \ref{fig:FeatureCombinationAndPCA} shows the purity of K-means clustering for different numbers of principal components for the three different data sets. PCA did not have a significant effect on the clustering performance. Except for data set 2, where the clustering performance decreased when using only a single principal component. The same applies to GMM clustering (data not shown).

\begin{figure}[htb]
    \centering
    \subfloat[\centering]{{\includegraphics[width=0.499\textwidth]{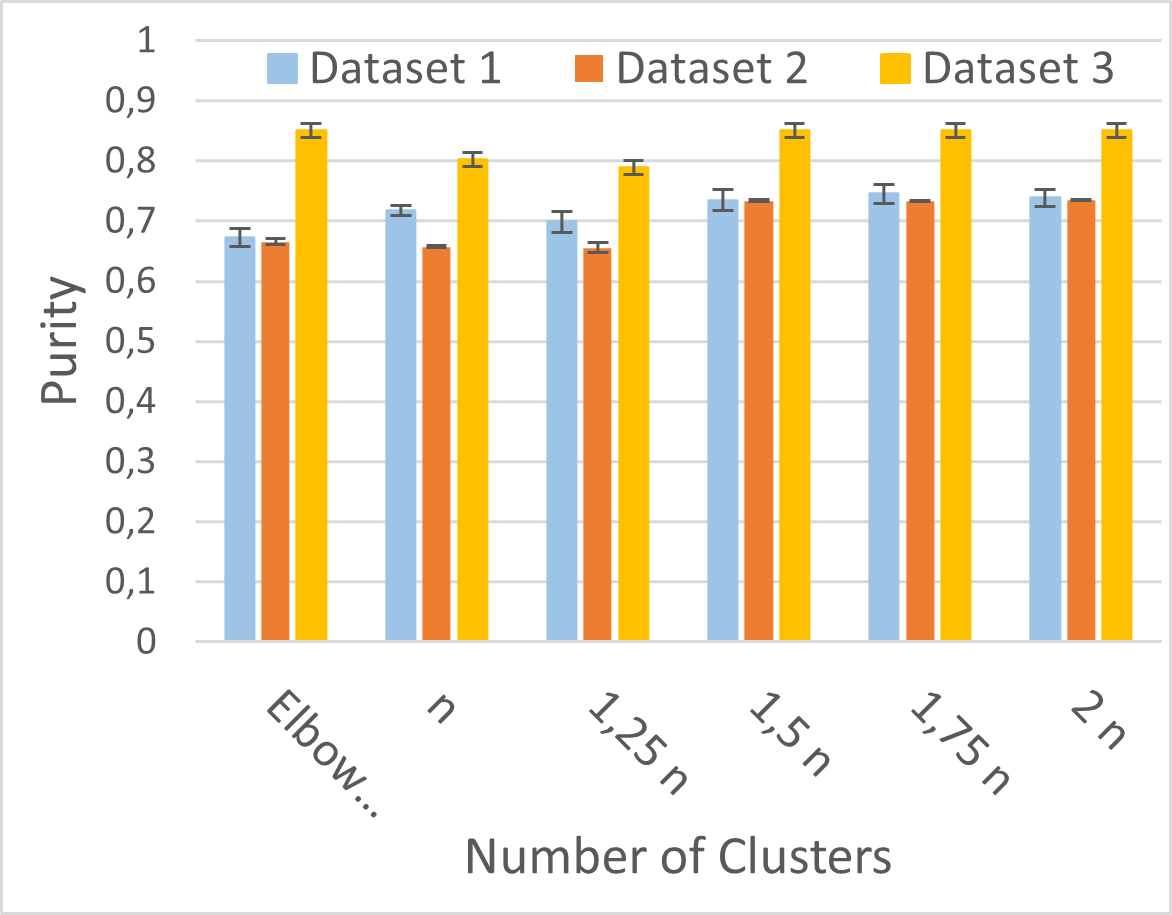}}}
    \subfloat[\centering]{{\includegraphics[width=0.501\textwidth]{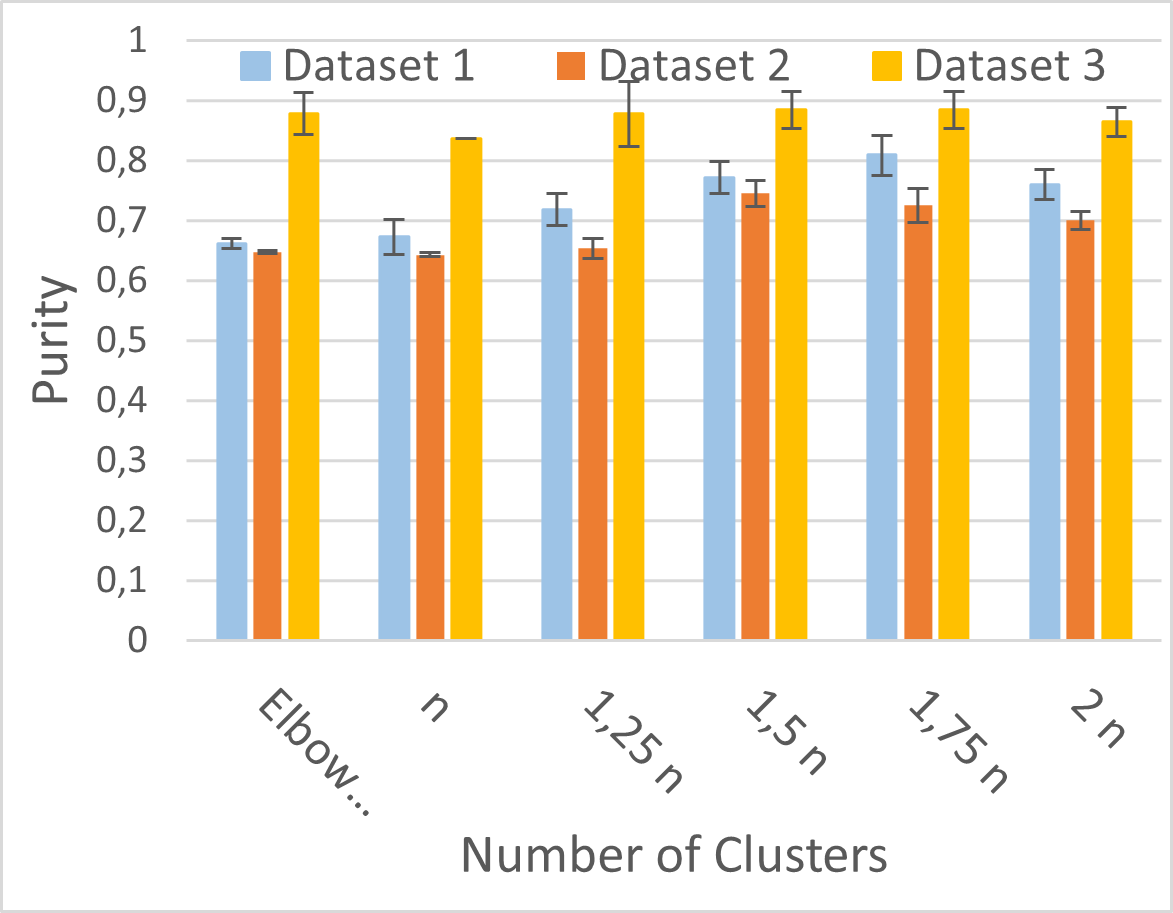}}}
    \caption{K-means clustering (a) and GMM clustering (b) purity for different specified numbers of clusters for different data sets}
    \label{fig:Overclustering}
\end{figure}

\textbf{Q5} Fig. \ref{fig:Overclustering} shows the purity of K-means clustering for different specified numbers of clusters for the three different data sets, as well as the purity for the number of clusters evaluated using the elbow method. The clustering performance did not change significantly for \(1.25n\), but significantly increased for \(1.5n\). A further increase in the specified number of clusters had no effect on the clustering performance. Fig. \ref{fig:Overclustering} shows the purity of GMM clustering for different specified numbers of clusters for the three different data sets, as well as the purity for the number of clusters evaluated using the elbow method. Clustering performance significantly increased for an increasing specified number of clusters, until \(2n\), where it slightly decreased.

\section{Discussion}
The high variance in purity shows that there is no general feature that performs best for an arbitrary data set, even though there are some trends for these specific data sets. Averaging (Mean, Median) and variance-based features (Standard Deviation, Interquartile Range) performed significantly better than shape-based features (Skewness, Kurtosis), even though these are frequently used in the literature \cite{Ruiz2014,Obuchowski2016,Dhamande2018}. This may also reflect on other shape-based features like the Crest Factor.

Even though OPTICS has proven useful for clustering vibration data in literature \cite{Hotait2021}, it clearly was not suited for the task of clustering these vibration data sets, as can be seen in Fig. \ref{fig:FraunhoferPlotsCombined}. Most of the data was labeled as noise by OPTICS. This could be a result of high variance and low class separability in the feature space of this data, which can be seen in Fig. \ref{fig:FraunhoferPlotsCombined}. It remains unclear if a more extensive optimization of OPTICS would have led to better results. It can be assumed that OPTICS would have needed specific optimization for every different dataset, which has not been done in this case. Therefore, OPTICS was discarded for the remaining three experiments.

Even though feature combinations for clustering are common practice in literature \cite{Ruiz2014,Obuchowski2016,Dhamande2018}, there was no significant performance improvement. All variations in performance can most likely be traced back to the intrinsic randomness of K-means and GMM.
PCA also did not result in any significant performance improvements. It is to note that even a single principal component seems to suffice for clustering, since it only resulted in a slight performance decrease.

As expected, increasing the number of clusters resulted in higher purity. It is to note that for K-means, an increase beyond \(1.5n\) did not result in any significant performance improvement. For GMM, a specified number of clusters as high as \(2n\) leads to a slight performance decrease. This could be a result of the GMM algorithm not being able to locate any more distinct Gaussian distributions in the data.

\section{Conclusion and Future Work}
In this work, we presented an extensive comparison of the clustering algorithms K-means clustering, OPTICS, and Gaussian mixture model clustering (GMM) using statistical features extracted from the time and frequency domains of three different vibration data sets. Furthermore, we investigated the influence of feature combinations, feature selection using principal component analysis (PCA), and the specified number of clusters on the performance of the clustering algorithms.

Our results showed that averaging (Mean, Median) and variance-based features (Standard Deviation, Interquartile Range) performed significantly better than shape-based features (Skewness, Kurtosis). In addition, K-means outperformed GMM slightly for these data sets, whereas OPTICS performed significantly worse than the other two algorithms. We were also able to show that feature combination as well as PCA feature selection did not result in any significant performance improvements. The performance of K-means increased significantly for a specified number of clusters of \(1.5\) times the number of conditions, but did not continue to increase with an increasing number of clusters. GMM's performance increased continuously until \(2\) times the number of conditions, when it began to decline.

A limitation of our study is that only three different data sets were used, and only three tests per experimental setting were run. This leads to uncertain conclusions about the generalizability of our results for arbitrary vibration data sets. Furthermore, this comparison is also limited to three specific clustering algorithms. Both limitations may be investigated in future studies.

\bibliographystyle{splncs04}
\bibliography{main}
\end{document}